\documentclass[letterpaper]{article} % DO NOT CHANGE THIS
\usepackage[submission]{aaai24}  % DO NOT CHANGE THIS
\usepackage{times}  % DO NOT CHANGE THIS
\usepackage{helvet}  % DO NOT CHANGE THIS
\usepackage{courier}  % DO NOT CHANGE THIS
\usepackage[hyphens]{url}  % DO NOT CHANGE THIS
\usepackage{graphicx} % DO NOT CHANGE THIS
\usepackage{natbib}  % DO NOT CHANGE THIS AND DO NOT ADD ANY OPTIONS TO IT
\usepackage{caption} % DO NOT CHANGE THIS AND DO NOT ADD ANY OPTIONS TO IT
\usepackage{algorithm}
\usepackage{algorithmic}
\usepackage{amsfonts,amssymb}
\usepackage{multirow}
\usepackage{array}

\usepackage{newfloat}
\usepackage{listings}
\DeclareCaptionStyle{ruled}{labelfont=normalfont,labelsep=colon,strut=off} % DO NOT CHANGE THIS
\floatstyle{ruled}
\newfloat{listing}{tb}{lst}{}
\floatname{listing}{Listing}
\title{centroIDA: Cross-Domain Class Discrepancy Minimization Based on Accumulative Class-Centroids for Imbalanced Domain Adaptation}
\author{
    Written by AAAI Press Staff\textsuperscript{\rm 1}\thanks{With help from the AAAI Publications Committee.}\\
    AAAI Style Contributions by Pater Patel Schneider,
    Sunil Issar,\\
    J. Scott Penberthy,
    George Ferguson,
    Hans Guesgen,
    Francisco Cruz\equalcontrib,
    Marc Pujol-Gonzalez\equalcontrib
}
\affiliations{
    \textsuperscript{\rm 1}Association for the Advancement of Artificial Intelligence\\
    1900 Embarcadero Road, Suite 101\\
    Palo Alto, California 94303-3310 USA\\
    proceedings-questions@aaai.org
}

\usepackage{bibentry}
\begin{document}

\maketitle

\begin{abstract}

Unsupervised Domain Adaptation (UDA) approaches address the covariate shift problem by minimizing the distribution discrepancy between the source and target domains, assuming that the label distribution is invariant across domains. However, in the imbalanced domain adaptation (IDA) scenario, covariate and long-tailed label shifts both exist across domains. To tackle the IDA problem, some current research focus on minimizing the distribution discrepancies of each corresponding class between source and target domains. Such methods rely much on the reliable pseudo labels' selection and the feature distributions estimation for target domain, and the minority classes with limited numbers makes the estimations more uncertainty, which influences the model's performance. In this paper, we propose a cross-domain class discrepancy minimization method based on accumulative class-centroids for IDA (centroIDA). Firstly, class-based re-sampling strategy is used to obtain an unbiased classifier on source domain. Secondly, the accumulative class-centroids alignment loss is proposed for iterative class-centroids alignment across domains. Finally, class-wise feature alignment loss is used to optimize the feature representation for a robust classification boundary. A series of experiments have proved that our method outperforms other SOTA methods on IDA problem, especially with the increasing degree of label shift.
\end{abstract}

\section{Introduction}

Unsupervised Domain Adaptation (UDA) aims to transfer knowledge from labeled source domain to unlabeled target domain. The existing UDA approaches usually are based on the distance measurement strategy \cite{long2015learning} and the adversarial learning strategy \cite{ganin2016domain}. The former derives a set of methods \cite{sun2016return,long2017deep,liu2020importance} devoted to finding appropriate discrepancy measures to reduce cross-domain distribution differences, whereas the latter derives a set of methods \cite{tzeng2017adversarial,long2018conditional,pei2018multi,chen2019transferability} that align cross-domain marginal distributions by adversarial learning between feature generators and domain discriminators. They mainly concentrate on resolving covariate shifts (feature shifts) across two domains, i.e., $p(x) \neq q(x)$ or $p(x|y) \neq q(x|y)$, where $x$ and $y$ represent the input data and labels, and $p$ and $q$ represent the distributions of the source and target domains, respectively. There is a potential assumption that classes are roughly balanced in each domain and label distribution is similar across domains.

\begin{figure*}[t]
\centering
\includegraphics[width=0.8\textwidth]{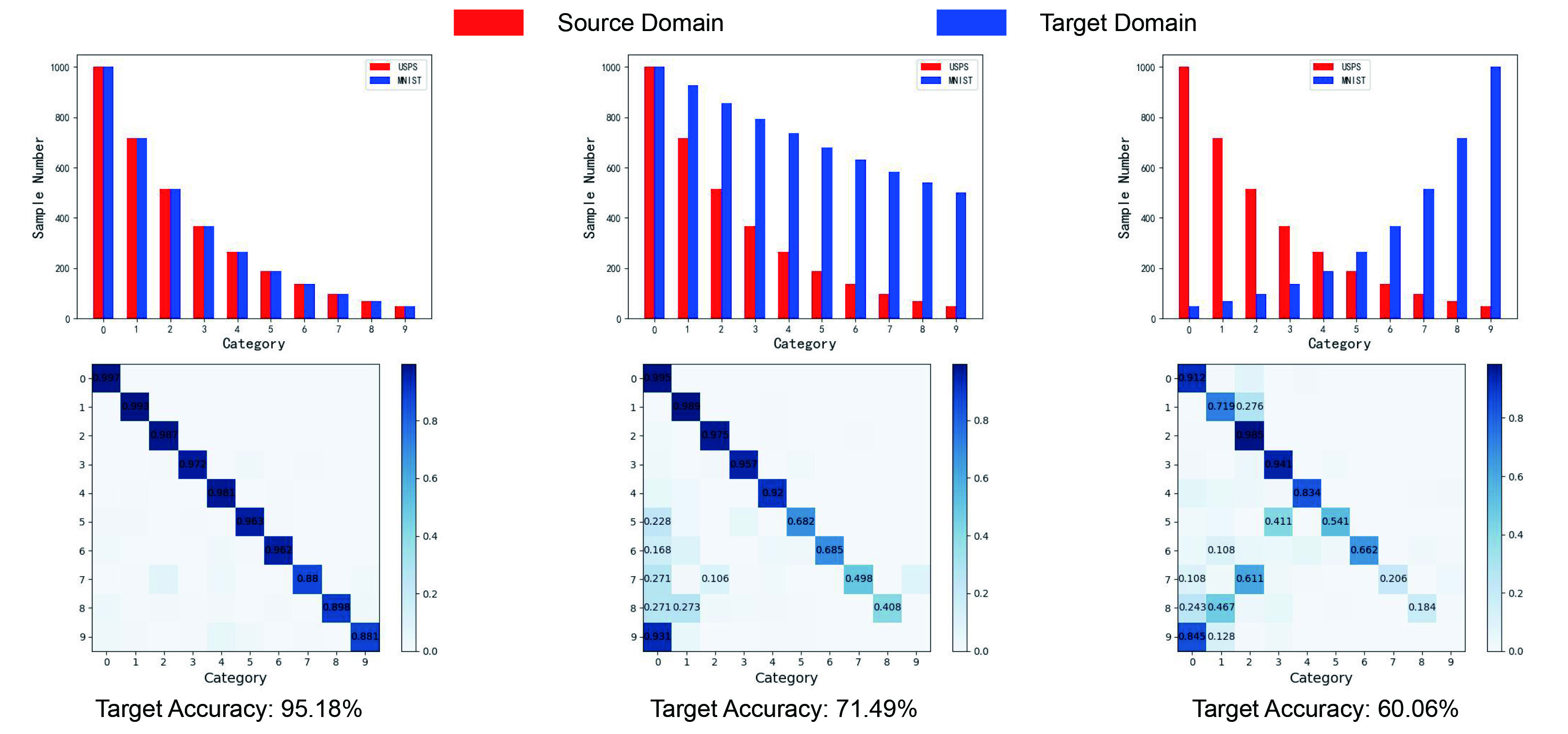} % Reduce the figure size so that it is slightly narrower than the column. Don't use precise values for figure width.This setup will avoid overfull boxes.
\caption{The effect of label shift on DANN. The upper figures depict the sample quantity of the datasets USPS and MNIST, and the lower figures depict the class confusion matrix obtained by the DANN  under the corresponding samples.}
\label{fig1}
\end{figure*}

However, the above assumption is not applicable to the problem of imbalanced domain adaptation (IDA) \cite{yang2021advancing}. IDA is a more realistic scenario in which the source and target domains have diversities of long-tailed distributions, which leads to label shifts. Fig. \ref{fig1} depicts three IDA scenarios: without label shift, moderate label shift, and severe label shift. Obviously, since the target's label is unknown, it is difficult to guarantee that both source and target domains follow a same distribution. Therefore, without label shift is an absolute idealization, and we will focus on IDA with label shift. Formally, in IDA, we assume $p(y|x)=q(y|x)$, but there are both covariate shifts $p(x) \neq q(x)$ and label shifts $p(y) \neq q(y)$ between the two domains.

% \subsection{Illustrations and  Figures}

\begin{figure}[H]
\centering
\includegraphics[width=0.9\columnwidth]{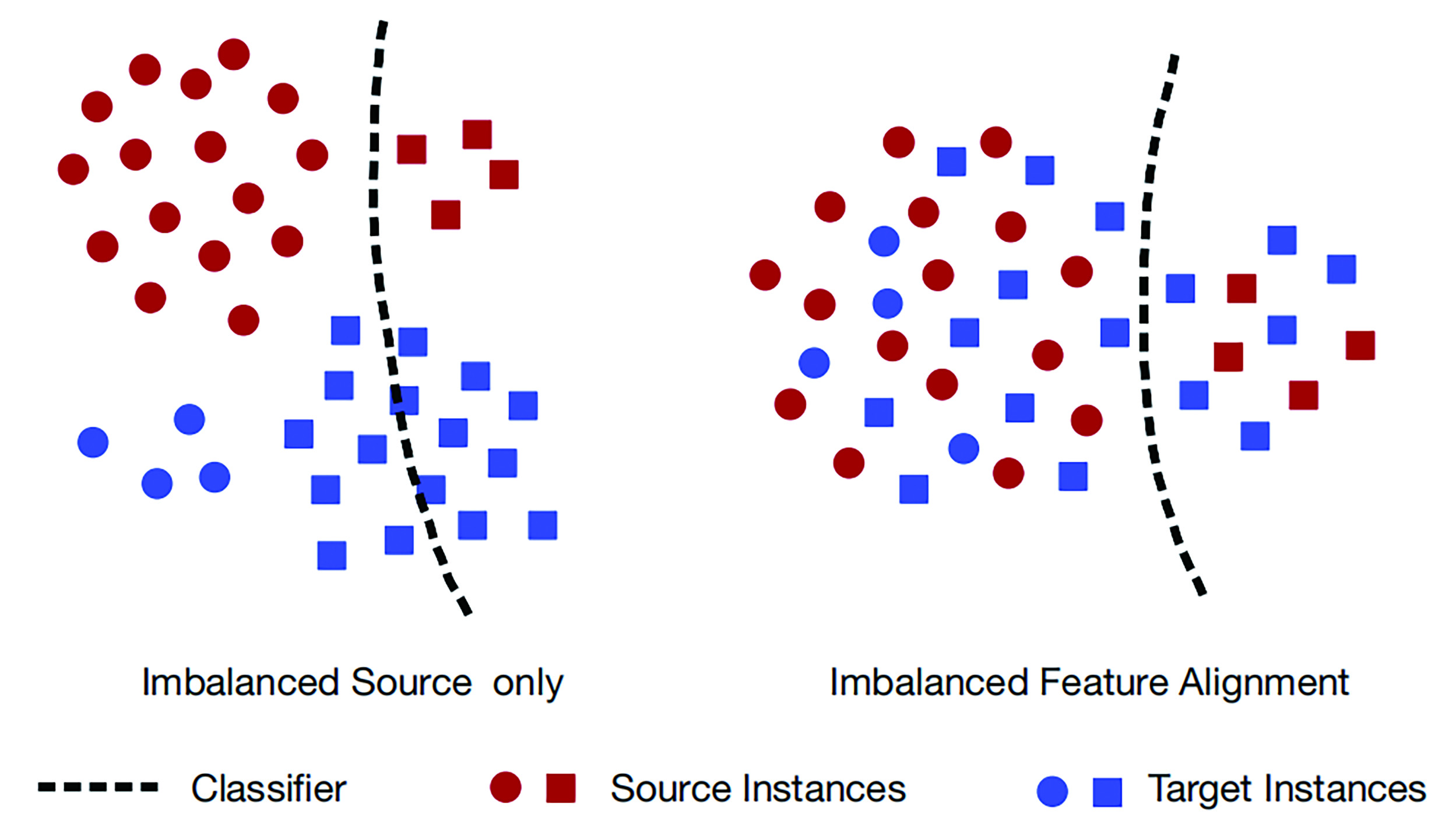} % Reduce the figure size so that it is slightly narrower than the column. Don't use precise values for figure width.This setup will avoid overfull boxes.
\caption{Incorrect feature alignment under label shift.}
\label{fig2}
\end{figure}

% 说明在IDA下导致的问题，通过实验等分析，详细说明三幅图中IDA问题场景下对传统模型造成的影响。
The IDA problem will result in model crashes and even negative transfer problem for most existing UDA methods. For example, we conducted an experiment to demonstrate the effect of label shift on the DANN model in three IDA scenarios, as shown in Fig. \ref{fig1}. The figure depicts the DANN's class confusion matrix with the corresponding data distribution. It can be seen that (\romannumeral1) when there is no label shift in two domains, DANN's performance is still good, even if the data are imbalanced, the accuracy of target minority classes will be slightly reduced owing to limited numbers; (\romannumeral2) When there is a moderate label shift in two domains, the accuracy of target minority classes drops significantly, and the most are incorrectly classified as majority classes; (\romannumeral3) When there is a significant label shift in two domains, the accuracy of the target classes corresponding to the source minority classes drops dramatically. In this case, some samples are incorrectly categorized as the source majority classes, while others are incorrectly categorized as other classes that are easily confused with them. This suggests that IDA's influence on the UDA model is mainly driven by the label shift under imbalanced distributions between source and target domains. DANN aligns target classes' features more with majority classes of source domain or easily confused classes in IDA, resulting in negative transfer of these target classes. Fig. \ref{fig2} provides an explanation for feature incorrectly alignment phenomenon caused by IDA. It demonstrates that in the IDA scenario with significant label shift, the feature alignment technique overfits the distributions of the source and target domains, resulting in negative transfer.

To tackle the IDA problem, some current studies \cite{yang2021advancing,tan2020class} achieve feature alignment by minimizing the distance between the corresponding classes between source and target domains. 
They achieve conditional feature alignment by aligning the source and target prototypes with prototype network\cite{snell2017prototypical}
 However, such methods use a pre-trained source-only model to calculate source prototypes, and then align them with target prototypes that calculated by source-target model (trained by source and target data). It is obvious that the source prototypes will change with the model's variation and it will result in an inaccurate feature alignment between the two domains. In addition, some other algorithms \cite{wang2023titok,yang2021advancing} use cluster-based or contractive learning method for aligning conditional feature alignment, and these methods mainly rely on the reliability of the pseudo labels on the target domain during training. However, the class-imbalance in IDA scenario will influence the training of feature extractor and classifier, thus it will disturb pseudo label's estimation. These factors are somewhat ignored by these algorithms.

In this paper, we propose cross-domain class discrepancy minimization based on accumulative class-centroids for IDA (centroIDA). First, according to \citeauthor{kang2019decoupling} \shortcite{kang2019decoupling}, class-balanced re-sampling is used on the source domain for obtaining an unbiased classifier, and it will ensure the reliability of the pseudo labels on the target domain. Second, accumulative class-centroids alignment is proposed to better estimate the feature distribution of target domain. Accordingly, the source and  target domains achieve alignment by iteratively updating the current class-centroids  according to the feature distributions learnt from new batches. Finally, class-wise feature alignment is used to optimize feature representations and to learn a robust classification boundaries between domains, by reducing intra-class distances and increasing inter-class distances among samples.

The contributions of this article are as follows:

\begin{itemize}
    \item We propose centroIDA for the IDA problem, which is based on class-balanced re-sampling, accumulative class-centroids alignment, and class-wise feature alignment. It ensures the reliability of pseudo label and target feature distribution estimation, which improves the knowledge transfer across domains with label shifts. 
    \item We originally propose accumulative class-centroids alignment, which uses both  the knowledge learnt from current batch's instances and previously trained instances to calculate the class-centroids for feature alignment. This provides a novel perspective for domain adaptation.
    \item A series of experiments proved that centroIDA outperforms other SOTA algorithms on the IDA problem, particularly as the degree of label shift increased.
\end{itemize}

\section{Related Work}

\subsubsection{Domain Adaptation With Covariate Shift}
is now the most concerning issue in UDA tasks, and research methods are broadly classified into three categories: statistic divergence alignment, adversarial training, and self-training. According to domain adaptation theory \cite{kouw2018introduction}, the target domain error's upper bound is constrained by the source domain error and the discrepancy between two domains. Consequently, many approaches seek suitable divergence metrics to minimize domain disparities within the feature space. Common measures include maximum mean discrepancy (MMD) \cite{jt2011domain}, Wasserstein distance \cite{liu2020importance}, correlation alignment (CORAL) \cite{sun2016return}, and contrastive domain discrepancy (CDD) \cite{kang2019contrastive}. For MDD-based methods, multiple kernel MDD (MK-MDD) \cite{long2015learning} and joint MMD (JMMD) \cite{long2017deep} have been proposed to improve the model's generalizability. CORAL defines the covariance of two domain features, and to quantify the difference in covariance, a series of distance explorations have been introduced \cite{sun2016deep,zhang2018unsupervised,wang2017deep,morerio2017minimal}. Inspired by the generative adversarial networks GANs, some methods use adversarial training to obtain domain invariant features, such as DANN \cite{ganin2016domain}, BSP \cite{chen2019transferability}, ADDA \cite{tzeng2017adversarial}, CDAN \cite{long2018conditional}, CADA \cite{zou2019consensus}, where BSP strikes a balance between transferability and discriminability.  Other methods use adversarial training to learn robust classification boundaries, such as TAT \cite{liu2019transferable} and MCD \cite{yang2020towards}. In addition to minimizing domain divergences, other strategies achieve self-training by utilizing unlabeled target domains \cite{mei2020instance,shin2020two,wei2020theoretical}. However, these approaches always suffer from unreliable pseudo labels of target domain. 

\subsubsection{Domain Adaptation With Label Shift}
aims to tackle the problem of different label distributions across domains. A series of methods \cite{lipton2018detecting,wu2019domain,azizzadenesheli2019regularized,alexandari2020maximum,9445225} predict and estimate the target label distribution to deal with the label shift between the source and target domains. However, they assume that the feature distribution is invariant and only focus on the label shift in domain adaptation. Some methods \cite{panareda2017open,yang2020heterogeneous,cao2018partial} have  looked into domain adaptation with incomplete overlap in cross-domain label spaces. They belong to a special label shift problem and will not be studied in this article. 

To overcome both label shift and covariate shift issues in IDA scenarios, 
some methods \cite{jiang2020implicit,prabhu2021sentry,liu2021adversarial} propose cross-domain alignment based on selective pseudo label strategies. However, selective strategies will neglect some information in the unlabeled samples on the target domain. While, some other methods use cluster-based pseudo-labels \cite{yang2021advancing} or class contractive knowledge \cite{wang2023titok} to overcome the imbalance sensitivity across domains. Nevertheless, how to guarantee the reliability of pseudo labels under imbalanced conditions is not well-designed.
Moreover, other methods \cite{tan2020class,yang2021advancing,9547057,9234105} achieve conditional feature distribution alignment by aligning the source and target prototypes based on prototype networks, i.e., aligning the centroids of the corresponding classes. However, the class prototypes on the source domain are calculated based on pre-trained source-only model, which ignores the transferring knowledge across domains, and will result in inaccurate alignment between source and target prototypes. In addition, prototype network will raise the computational costs due to the calculation complexity of entire samples, which is not acceptable for huge datasets. Thus, reliable target pseudo labels and accurate conditional feature distribution alignment should be considered in IDA, as well as complexity. 

% Thus, the class prototype would be better calculated during DA training with more accurate distribution estimation and lower complexity.
% Other methods \cite{tan2020class,yang2021advancing,9547057,9234105} are studied based on prototype networks to make the prototypes for each class in the source and target domains close in the embedding space, i.e., aligning the centroids of the corresponding classes. However, they minimize the distance between the source prototypes from the source only model and the target prototypes from the source-target model, ignoring the source prototypes' differences between the two models, resulting in inaccurate alignment, and the use of prototype networks raises computational costs. Therefore, in response to the unreliable pseudo labels and inaccurate target distribution estimation in IDA problems, we propose cross-domain class discrepancy minimization based on accumulative class-centroids. 

\section{Method}
\subsection{Methodology Overview}

\begin{figure*}[t]
\centering
\includegraphics[width=0.8\textwidth]{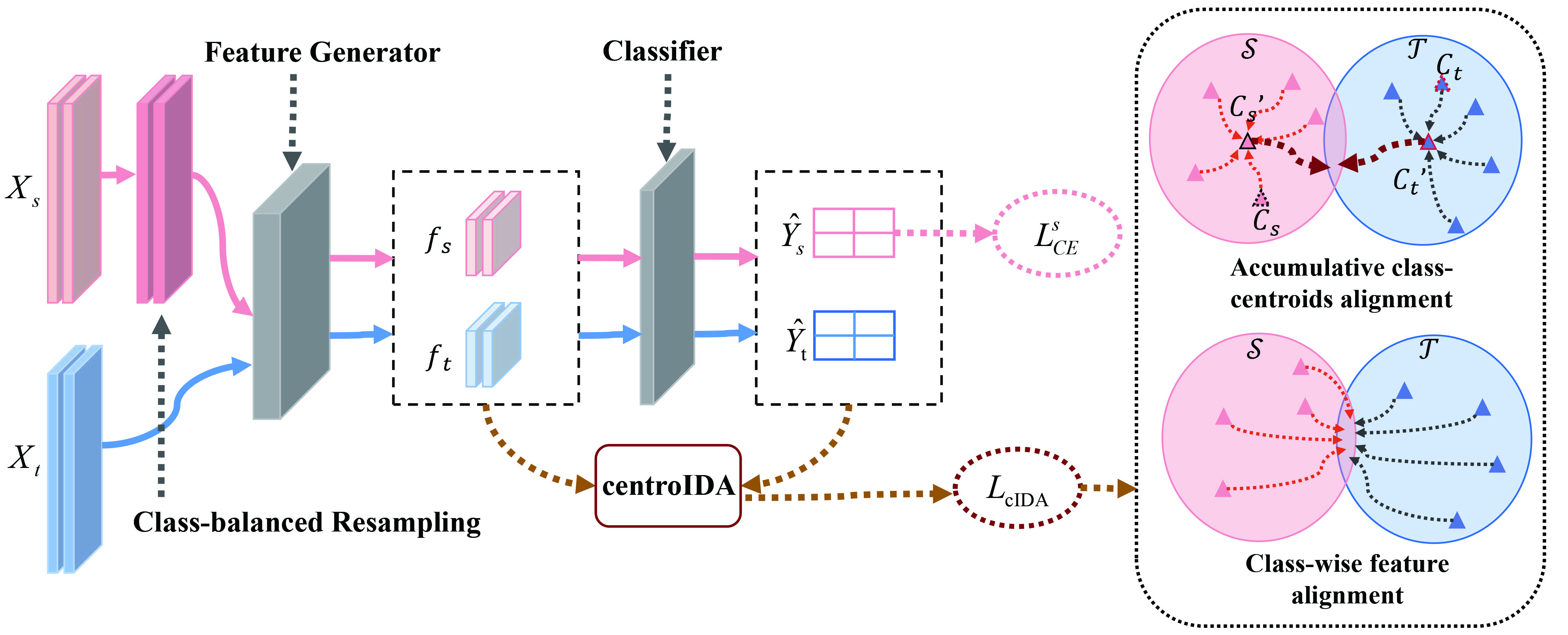} % Reduce the figure size so that it is slightly narrower than the column. Don't use precise values for figure width.This setup will avoid overfull boxes.
\caption{Network structure of centroIDA.}
\label{fig3}
\end{figure*}

In the IDA issue, we are given a source domain $\mathcal{S}=\{\mathcal{X}_i^s,\mathcal{Y}_i^s\} $ with $N_s$ labeled samples $\{(x_i^s,y_i^s)_{i=1}^{N_s}\}$ and a target domain $\mathcal{T}=\{\mathcal{X}_i^t\}$ with $N_t$ unlabeled samples $\{(x_i^t)_{i=1}^{N_t}\}$.
Assuming that the conditional label distribution of $\mathcal{S}$ and $\mathcal{T}$ is invariant $p(y|x)=q(y|x)$, but there are both covariate shifts $p(x)\neq q(x)$ and label shifts $p(y)\neq q(y)$. We aim to design a neural network that can transfer the knowledge  from $\mathcal{S}$ to $\mathcal{T}$ while minimizing the categorized risk on $\mathcal{T}$. Our network structure can be seen in Fig.\ref{fig3}

First, to overcome the negative effects of long-tailed distribution, class-balanced re-sampling is used \cite{kang2019decoupling} on $\mathcal{S}$ for guaranteeing the classifier's unbiased performance. Specifically, the sampling probability for each class is equal during the training, and for a K-classification problem, the probability of each class being sampled is $1/K$. Then, accumulative class-centroids alignment is proposed to establish alignment between the two domains in order to avoid the impact of label shift. Finally, the pseudo labels of $\mathcal{T}$ are corrected based on the closest distance to target class-centroids, and class-wise feature alignment is used to optimize the feature representation for a robust classification boundary. It is worth noting that we use class-balanced re-sampling rather than re-weighting \cite{ren2018learning} because re-sampling ensures equal participation of each class in training to update class-centroids, preventing the accumulative class-centroids for minority classes being empty for a period of time that making alignment impossible. 

\subsection{Accumulative Class-Centroids Alignment}
In this section, we calculate the accumulative class-centroids of $\mathcal{S}$ and $\mathcal{T}$ for iterative class-centroids alignment across domains. To avoid overconfident predictions, we use temperature rescaling \cite{guo2017calibration} to calculate the probability $P_{ij}$ of the sample $x_i$ being classified into category $y_j$ as:
\begin{equation} \label{eq:(1)}
    \widehat{Y}_{ij}=\frac{\exp {(Z_{ij}/T)}}{\sum_{k=1}^{|\mathcal{C}|} \exp {(Z_{ik}/T)}} ,
\end{equation}
where $Z_{ij}$ is the classifier's logical output, $|\mathcal{C}|$ denotes the number of classes and $T$ is the temperature hyperparameter, with $T=2$. Then, the maximum classification probability $P_i^{\max }$ of the sample $x_i$ can be expressed as:
\begin{equation} \label{eq:(2)}
    P_i^{\max }=\max \{ \widehat{Y}_{ij} | j=1,\cdots|\mathcal{C}| \} .
\end{equation}
Unlike class-centroids proposed by \citeauthor{tian2020domain} \shortcite{tian2020domain}, we believe that the importance of each instance is different, and that those with higher reliability should be given more attention.  For instance, in binary classification, the classification probabilities for two examples are $[0.8,0.2]$ and $[0.6,0.4]$, respectively. Obviously, the former is  more dependable. As a result, we compute the accumulative class-centroids using  ${P^{\max }}$ as an example's weight.

\subsubsection{Accumulative Class-Centroids}
Acquiring each class's centroid within a batch is usually impracticable, especially when the number of categories exceeds the batch size, because the training process is batch-based, making it difficult to gather all sample information at once. So some methods \cite{tan2020class,pan2019transferrable,yang2021advancing} rely on prototype networks to align source and target prototypes. But the source and target prototypes calculated in different model, resulting in incorrect alignment. Therefore, we propose accumulative class-centroids for alignment, where the source and target class-centroids are iteratively updated and calculated both from the source-target model.

We use ${f(x_i)}$ to represent the feature point of the example $x_i$ obtained by the feature generator, $P_i^{\max}$ to represent its weight, and $f(x)$ and $P^{\max} $ together decide the class-centroids. During the training phase, the class-centroids of the source and target domains are initialized to $\mathbf{0}$. And in each iteration, each class's centroids will be updated with the current batch instances. We will acquire the class-centroids of all instances when all examples have been learned. Because the training is conducted in batches, the centroids obtained during the process are the centroids of the examples that have been trained, rather than the centroids of all examples. Therefore, we refer to them as accumulative class-centroids and keep track of them, as well as their accompanying accumulative weights, at each iteration. The accumulative class-centroids and accumulative weight of the k-th class are represented by $C_k$ and $P_k^c$, respectively. Therefore, the new round $C_k'$ can be written as:
\begin{equation} \label{eq:(3)}
    C_k'=\frac{C_k\cdot P_k^c + \sum_{i=1,\cdots B,L_i=k}f(x_i)\cdot P_i^{\max}}{P_k^c+\sum_{i=1,\cdots B,L_i=k} P_i^{\max}},
\end{equation}
% \begin{equation} \label{eq:(3)}
%     C_k'=\frac{C_k\cdot P_k^c + \sum_{L_i=k}f(x_i)\cdot P_i^{\max}}{P_k^c+\sum_{L_i=k} P_i^{\max}},
% \end{equation}
where $B$ is the batch size and $L_i$ represents the pseudo label based on classifier of example $x_i$. The current accumulative class-weight is updated to:
\begin{equation} \label{eq:(4)}
    {P_k^c}'=P_k^c+\sum_{i=1,\cdots B,L_i=k} P_i^{\max}.
\end{equation}
% \begin{equation} \label{eq:(4)}
%     {P_k^c}'=P_k^c+\sum_{L_i=k} P_i^{\max}.
% \end{equation}
According to Eq.(\ref{eq:(3)}), we obtain accumulative class-centroids of the source domain $C_k^s$ and accumulative classcentroids of the target domain $C_k^t$. Then, our accumulative class-centroids alignment loss $loss\_ c$ is expressed as:
\begin{equation} \label{eq:(5)}
    loss\_ c =\frac{|\mathcal{C}| \sum_{k=1}^{|\mathcal{C}|} {\| C_k^s-C_k^t \| }}{ \sum_{i,j=1,\cdots |\mathcal{C}|}{\| C_i^s-C_j^t\|}},
\end{equation}
% \begin{equation}
%     loss\_ c=\frac{\frac{1}{|C|}}{}
% \end{equation}
where $\| \cdot\|$ is the L2-norm. From Eq.(\ref{eq:(5)}), It is evident that this promote the separation between distinct classes while also decreasing the distance between centroids of the same class in the two domains. But the features generated by this centroid-based  method is still scattered, stricter alignment criteria are still required to further encourage  learning  representations of the target domain.

\subsection{Class-Wise Feature Alignment}

% In this section, we look at solutions to further reducing the discrepancies between  corresponding clusters cross-domains.

% The distance between feature points that instance mapping in the feature space reflects their differences. We aim to (\romannumeral1) obtain a robust classification boundary by minimizing a distance between instances of the same class and maximizing the separation between different classes; and (\romannumeral2) achieve knowledge sharing in class discrimination by getting closer the sample features between the corresponding class clusters in two domains. 

Class-wise feature alignment is proposed in this section to optimize the feature representation for a robust classification boundary. To reduce classification risk in the target domain, we intend to (\romannumeral1) reduce intra-class differences and widen inter-class intervals; and (\romannumeral2) align the features of the corresponding classes in the two domains. Then we calculate the distance between any two examples from $\mathcal{S}$ and $\mathcal{T}$, and it is as follows:
\begin{equation}  \label{eq:(6)}
    d_{ij}=\| f(x_i^s)-f(x_j^t)\|,
\end{equation}
where $d_{ij}$ represents the distance between the example $x_i^s$ from $\mathcal{S}$ and the example $x_j^t$ from $\mathcal{T}$ in the feature space. It is crucial to guarantee that the target pseudo labels are relatively trustworthy in order to produce a more reliable class-wise feature alignment. The classifier's classification findings are more appropriate for the source domain, thus labeling the target sample by the classifier may be inaccurate. Therefore, the target pseudo labels are corrected based on the accumulative class-centroids.
We label $x^t$ as $\widehat{L}^t$ based on the closest distance to the accumulative class-centroids, and label $x^s$ as $L^s$ by the classifier. When the two labels are the same, $d$ is expected to be as tiny as possible; otherwise, $d$ would be larger. In addition, we note that even if the $P^{\max}$ of two instances is equal, their contribution to reducing class feature differences should not be identical. For example, based on the classifier's logical output, the classification probabilities for instances $a$ and $b$ are [0.5, 0.4, 0.1, 0.0], and [0.5, 0.1, 0.2, 0.2], respectively. To prevent class confusion, the weight of $a$ should be greater than that of $b$. Inspired by \cite{jin2020minimum}, we use entropy function $H(p)=-\mathbb{E}_p \log p$ for measuring, and the entropy $H(\widehat{\mathrm{y} }_{i\cdot})$ of the instance $x_i$ can be defined as: 
\begin{equation} \label{eq:(7)}
 H(\widehat{\mathrm{y} }_{i\cdot})=-\sum_{j=1}^{|\mathcal{C}|} \widehat{Y}_{ij} \log \widehat{Y}_{ij}.
\end{equation}
Then, the weight of the example $x_i$ is updated as:
\begin{equation} \label{eq:(8)}
    W_i= \frac{B(1+\exp(-H(\widehat{\mathrm{y} }_{i\cdot})))}{\sum_{k=1}^B(1+\exp(-H(\widehat{\mathrm{y} }_{k\cdot}))) } \cdot P_i^{\max}.
\end{equation}
The discrepancies between the class-wise feature of two domains can be represented by their average distance $d_{same}$:
\begin{equation} \label{eq:(9)}
    d_{same}=\frac{\sum_{i,j=1,\cdots B,L_i^s = \widehat{L}_j^t}{\sqrt{W_i^sW_j^t}\cdot d_{ij}}}{\sum_{i,j=1,\cdots B,L_i^s = \widehat{L}_j^t}{\sqrt{W_i^sW_j^t}}}.
\end{equation}
% \begin{equation} \label{eq:(9)}
%     d_{same}=\frac{1}{n_1}\sum_{i=1}^B\sum_{j=1,\cdots B,L_i^s = \widehat{L}_j^t}{\sqrt{W_i^sW_j^t}\cdot d_{ij}}
% \end{equation}
In the Eq.(\ref{eq:(9)}), $L_i^s$ denotes the pseudo label of source instance $x_i^s$ by the classifier, and $L_j^t$ represents the pseudo label of target instance $x_j^t$ based on the closest distance to target accumulative class-centroids.
Similarly, we use average distance between different class feature in two domains to represent inter-class divergences. The inter-class discrepancy $d_{diff}$ is given by:
\begin{equation} \label{eq:(10)}
    d_{diff}=\frac{\sum_{i=1}^B\sum_{j=1,\cdots B,L_i^s \neq \widehat{L}_j^t}{\sqrt{W_i^sW_j^t}\cdot d_{ij}}}{\sum_{i,j=1,\cdots B,L_i^s \neq \widehat{L}_j^t}{\sqrt{W_i^sW_j^t}}}.
\end{equation}
Then $loss\_d$ is defined as $d_{same}$ divided by $d_{diff}$, since it encourages class-wise feature alignment between the two domains, keeping dissimilar class feature away and reinforcing the classification boundary. The formula is given by:
\begin{equation} \label{eq:(11)}
    loss\_d=d_{same}/d_{diff}.
\end{equation}

% \subsection{Overall Objective}

\begin{algorithm}[tb]
\caption{centroIDA}
\label{algorithm1}
% \textbf{Input}: Labeled source examples $(x^s,y^s)$, unlabeled target data $(x^t)$\\
% \textbf{Parameter}: Optional list of parameters\\
% \textbf{Output}: Optimal network model parameters $\theta$ \\
\begin{algorithmic}[1] %[1] enables line numbers
\STATE \textbf{Input}: Labeled source examples $(x^s,y^s)$, unlabeled target data $(x^t)$,parameters $\lambda$ and $\gamma$\\
\FOR{each epoch}
\STATE Source class-balanced re-sampling.
\STATE Initialize class-centroids $C^s$ and $C^t$ to $\mathbf{0}$, accumulative weight $P^s$ and $P^t$ to 0.
\FOR{each iteration}
\STATE Update $C^s$ and $C^t$ via Eq.(\ref{eq:(3)}) and update $P^s$ and $P^t$ via Eq.(\ref{eq:(4)})
\STATE Calculate accumulative class-centroids alignment loss $loss\_c$ in Eq.(\ref{eq:(5)})
\STATE Calculate class-wise feature alignment loss $loss\_d$ in Eq.(\ref{eq:(11)})
\STATE Update network model parameters $\theta$ by minimizing all losses in Eq.(\ref{eq:(13)})
\ENDFOR
\ENDFOR
\STATE \textbf{Output}: Optimal network model parameters $\theta$ \\

\end{algorithmic}
\end{algorithm}

In summary, our centroIDA model takes into account accumulative class-centroids alignment, class-wise feature alignment, and source class-balanced re-sampling. The loss $L_{cIDA}$ is defined as:
\begin{equation} \label{eq:(12)}
    L_{cIDA}=\lambda loss\_c+\gamma loss\_d,
\end{equation}
where $\lambda$ and $\gamma$ are hyper-parameters that balance the contribution of two losses. Besides, we use the cross-entropy loss $L_{CE}^s$ to minimize classification error in the source domain. Then, our general optimization objective is formulated as:
\begin{equation} \label{eq:(13)}
    \min\limits_{\theta} L_{CE}^s+L_{cIDA}.
\end{equation}
The overall goal is optimized through backpropagation. The training process of centroIDA is depicted in Algorithm \ref{algorithm1}.

\section{Experiments}
To investigate the efficacy of centroIDA on IDA problem, sampling protocol with a imbalanced ratio $p$ is proposed to conduct label shifts on existing UDA datasets. We evaluated centroIDA on two datasets, Office-Home \cite{venkateswara2017deep} and Domain-Net \cite{peng2019moment}, and then provide analysis of centroIDA, including different degrees of label shift, hyper-parameter analysis, ablation study and complexity analysis.

\subsection{Setup}
\subsubsection{Sampling Protocol}
A dataset with $N_c$ categories is ranked in descending order based on instance counts, with $N_{\max}$ being the classes' maximum number and $p(p \leq 1)$ being the imbalanced ratio. Assuming the class with label $i$ has a sorted order $j$, the samples' number in this class is $N_i \approx N_{\max} \cdot p^{j/(N_c-1)} $. This sampling strategy adheres to the datasets' original imbalance and allows to investigate the influence of varied degrees of label shift.

\subsubsection{Datasets}
\begin{figure*}[t]
\centering
\includegraphics[width=0.9\textwidth]{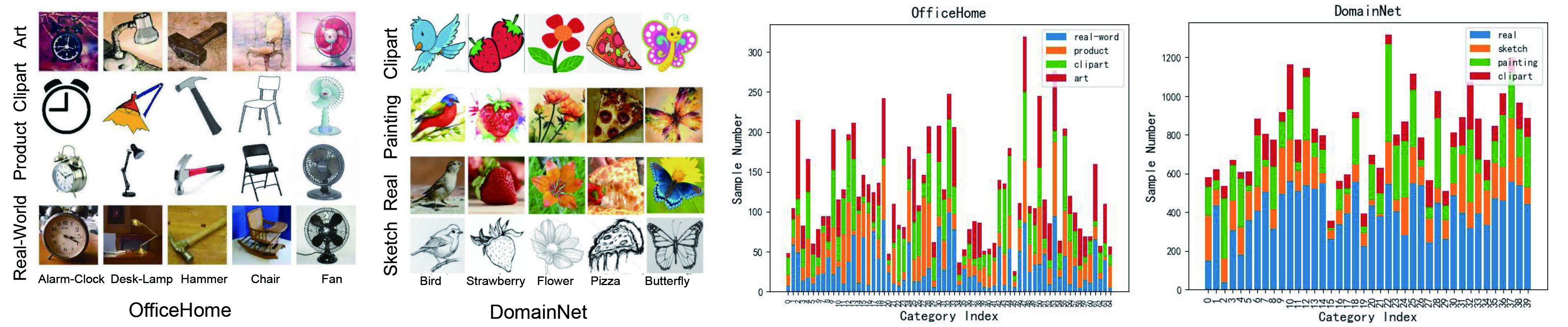} % Reduce the figure size so that it is slightly narrower than the column. Don't use precise values for figure width.This setup will avoid overfull boxes.
\caption{Image examples and data distribution of OfficeHome and DomainNet.}
\label{fig4}
\end{figure*}

We evaluated our model on two datasets: 1) Office-Home, sub-sampled with imbalanced ratio  $p=0.05$ from its four distinctive domains (Art (Ar), Product (Pr), Clipart (Cl), and RealWorld (Rw)) according to the sampling protocol; 2) DomainNet, forty common categories are selected from four fields (Real (R), Clipart (C), Painting (P), Sketch (S)), since there are serious label shift between them, it has not been sub-sampled. Fig. \ref{fig4} shows the images and data distribution of OfficeHome and DomainNet.

\subsubsection{Baseline Methods}
We follow the UDA evaluation procedure and conduct all experiments using the Pytorch framework and resnet50 as the backbone. We compare our method with currently advanced domain adaptation methods: DAN \cite{long2015learning}, DANN \cite{ganin2016domain}, CDAN \cite{long2018conditional}, BSP \cite{chen2019transferability}, MDD \cite{zhang2019bridging}, and TIToK \cite{wang2023titok}. DAN and MDD achieve alignment by decreasing the discrepancies between the two domains; DANN, CDAN, and BSP are based on adversarial networks; and TIToK is proposed to deal with bi-imbalanced UDA for label shift. All baselines are evaluated with their official codes and hyper-parameters. 

\subsubsection{Experimental Details}
We specify the following hyper-parameters for our model: $\lambda=5$, $\gamma=\frac{2}{\exp (-10\alpha)}-1$, where $\alpha$ varies linearly from 0 to 1. We set the bottleneck layer dimension to 256, batch size to 50, and use Stochastic Gradient Descent optimizer (SGD) with momentum 0.9. On Office-Home and Domain-Net, the classifier's initial Learning rate is set to 0.005 and 0.01 correspondingly, and its value is changed according to \cite{ganin2016domain}. We adopt the per-class mean accuracy \cite{dong2018imbalanced} as the evaluation indicator for the accuracy cannot appropriately reflect the performance in the case of imbalance data. Formally, the accuracy of the k-th class can be expressed as $Acc_k=n_{(k,k)}/N_k$, where $n_{(i,j)}$ represents the number of instances from class $i$ being labeled class $j$, $N_k$ represents the number of instances in the k-th class. Then the per-class mean accuracy can be expressed as: $Acc=\frac{1}{|\mathcal{C}|} \sum_1^{|\mathcal{C}|} Acc_k$, where $|\mathcal{C}|$ stands for the number of classes. In addition, the imbalanced data obtained from Office-Home according to the sampling protocol is only used for training, and we still use the original Office-Home data for testing, ensuring more reliable test results.

\subsection{Main Results}
Table \ref{table1} and Table \ref{table2} show the results of Office-Home and Domain-Net, respectively, with the top performance highlighted in bold. Some approaches, such as DANN and BSP, perform worse than the source only model on some IDA tasks since they only consider feature shifts, resulting in incorrect class feature alignment. Comepared with the performance on DomainNet, centroIDA has a better advantage on Office-Home than other baselines since we process it with a higher label shift according to the sampling protocol. It is worth noting that the classes' maximum number in the four Office-Home domains is 99, when we sample at $p=0.05$, the classes' minimum number is just 5, demonstrating that centroIDA performs better in more extreme label shifts. We also conduct experiments with different imbalanced ratios p, as shown in Table \ref{table3}, to explore centroIDA's ability to handle covariate shifts and resist label shifts. Obviously, in all circumstances, our method outperforms all baselines.

\begin{table*}[t]

% \resizebox{\textwidth}{!}{
% \renewcommand{\arraystretch}{1.2}
% \setlength{\tabcolsep}{0.6mm}
\centering
% \small
    
% \resizebox{.95\textwidth}{0.7in}{
% \begin{tabular}{l|m{0.7cm} m{0.7cm} m{0.7cm} m{0.7cm} m{0.7cm} m{0.7cm} m{0.7cm} m{0.7cm} m{0.7cm} m{0.7cm} m{0.7cm} m{0.7cm}|m{0.7cm} }
\begin{tabular}{@{}l@{}|@{ }c @{ }c @{ }c @{ }c @{ }c @{ }c @{ }c @{ }c @{ }c @{ }c @{ }c @{ }c| @{ }c@{} }
% \begin{tabular}{ 
%   | >{\raggedright\arraybackslash}X 
%    >{\centering\arraybackslash}X 
%    >{\centering\arraybackslash}X
%    >{\centering\arraybackslash}X
%    >{\centering\arraybackslash}X
%    }
    \hline

   \centering 
   Methods 
   & Cl \hspace{-0.8mm}$\to$\hspace{-0.8mm} Pr 
   & Pr \hspace{-0.8mm}$\to$\hspace{-0.8mm} Cl 
   & Ar \hspace{-0.8mm}$\to$\hspace{-0.8mm} Rw 
   & Rw \hspace{-0.8mm}$\to$\hspace{-0.8mm} Ar 
   & Rw \hspace{-0.8mm}$\to$\hspace{-0.8mm} Pr 
   & Pr \hspace{-0.8mm}$\to$\hspace{-0.8mm} Rw 
   & Ar \hspace{-0.8mm}$\to$\hspace{-0.8mm} Cl 
   & Cl \hspace{-0.8mm}$\to$\hspace{-0.8mm} Ar 
   & Ar \hspace{-0.8mm}$\to$\hspace{-0.8mm} Pr 
   & Pr \hspace{-0.8mm}$\to$\hspace{-0.8mm} Ar 
   & Cl \hspace{-0.8mm}$\to$\hspace{-0.8mm} Rw 
   & Rw \hspace{-0.8mm}$\to$\hspace{-0.8mm} Cl 
   & Mean \\

    \hline
     Source Only & 43.0 & 31.5 & 66.5 & 52.9 & 63.1 & 62.6 & 36.8 & 35.4 & 53.1 & 42.5 & 49.2 & 35.0 & 47.6 \\

     \hline
     DAN & 53.5 & 35.1 & 67.4 & 55.1 & 69.9 & 65.1 & 42.4 & 42.1 & 60.5 & 43.5 & 56.8 & 42.3 & 52.8 \\

     DANN & 45.1 & 37.1 & 64.4 & 53.7 & 63.0& 62.0 & 42.0 & 38.0 & 53.7 & 42.3 & 48.9 & 42.4 & 49.4 \\

     BSP & 49.2 & 39.7 & 65.2 & 54.6 & 65.6 & 62.1 & 45.0 & 40.5 & 54.7 & 44.1 & 51.5 & 44.2 & 51.4 \\

     CDAN & 48.5 & 37.8 & 64.7 & 53.1 & 67.8 & 63.1 & 41.2 & 36.6 & 57.1 & 41.4 & 49.6 & 43.3 & 50.4 \\

     MDD & 55.0 & 42.4 & 69.0 & 56.1 & 70.9 & 65.4 & 47.6 & 43.7 & 60.6 & 45.3 & 56.4 & 46.4 & 54.9 \\

     TIToK & 56.5 & 40.5 & 70.0 & 60.0 & 70.5 & 67.9 & 43.7 & 39.6 & 61.7 & 49.8 & 56.6 & 44.9 & 55.1 \\

     \hline
     centroIDA & \textbf{62.9} & \textbf{46.2} & \textbf{70.3} & \textbf{62.0} & \textbf{74.9} & \textbf{72.2} & \textbf{49.8} & \textbf{51.8} & \textbf{67.4} & \textbf{57.3} & \textbf{63.1} & \textbf{50.9}  & \textbf{60.7}\\
     \hline

\end{tabular}
% }
\caption{Per-class mean accuracy (\%) on OfficeHome with imbalanced ratio $p=0.05$}
\label{table1}
\end{table*}

\begin{table*}[t]
\centering
% \small 

\begin{tabular}{@{}l@{}| @{\ \ }c@{\ \ } @{\ \ }c@{\ \ } @{\ \ }c@{\ \ } @{\ \ }c@{\ \ } @{\ \ }c@{\ \ } @{\ \ }c@{\ \ } @{\ \ }c@{\ \ } @{\ \ }c@{\ \ } @{\ \ }c@{\ \ } @{\ \ }c@{\ \ } @{\ \ }c@{\ \ } @{\ \ }c@{\ \ }| @{\ \ }c@{} }
    \hline

    Methods & R\hspace{-0.8mm}$\to$\hspace{-0.8mm} C & R \hspace{-0.8mm}$\to$\hspace{-0.8mm} P & C \hspace{-0.8mm}$\to$\hspace{-0.8mm} R & C \hspace{-0.8mm}$\to$\hspace{-0.8mm} P & P \hspace{-0.8mm}$\to$\hspace{-0.8mm} R & P \hspace{-0.8mm}$\to$\hspace{-0.8mm} C & R \hspace{-0.8mm}$\to$\hspace{-0.8mm} S & S \hspace{-0.8mm}$\to$\hspace{-0.8mm} R & C \hspace{-0.8mm}$\to$\hspace{-0.8mm} S & S \hspace{-0.8mm}$\to$\hspace{-0.8mm} C & P \hspace{-0.8mm}$\to$\hspace{-0.8mm} S & S \hspace{-0.8mm}$\to$\hspace{-0.8mm} P & Mean \\

    \hline
     Source Only & 60.2 & 65.1 & 72.0 & 51.8 & 81.5 & 55.9 & 46.6 & 73.9 & 49.6 & 60.3 & 51.6 & 53.5 & 60.2 \\

     \hline
     DAN & 69.1 & 71.8 & 80.5 & 61.3 & 84.7 & 64.9 & 62.8 & 80.8 & 59.5 & 71.1 & 65.2 & 67.4 & 70.0\\

     DANN & 58.6 & 61.7 & 71.1 & 55.2 & 79.3 & 59.7 & 70.7 & 80.4 & 65.2 & 72.5 & 62.4 & 65.5 & 66.9 \\

     BSP & 74.8 & 73.5 & 79.8 & 63.7 & 79.5 & 62.4 & 72.5 & 81.6 & 67.8 & 74.1 & 61.5 & 67.4 & 71.6\\

     CDAN & 76.9 & 74.9 & 87.1 & 67.3 & 82.5 & 63.9 & 69.7 & 83.1 & 67.8 & 76.0 & 66.8 & 68.9 & 73.7 \\

     MDD & 77.1 & 74.1 & 87.0 & 70.1 & 85.7 & 68.9 & 75.6 & 85.7 & 71.0 & \textbf{79.2} & 69.7 & 70.8  &  76.2   \\

     TIToK & 78.3 & 76.2 & 86.9 & 69.3 & \textbf{87.6} & 74.8 & 73.7 & 85.8 & 71.1 & 79.1 & 70.2 & 71.8 & 77.1   \\

     \hline
     centroIDA & \textbf{82.2} & \textbf{77.2} & \textbf{88.6} & \textbf{73.6} & 87.4 & \textbf{78.3} & \textbf{79.5} & \textbf{86.9} & \textbf{75.4} & 78.6 & \textbf{76.1} & \textbf{72.2} & \textbf{79.7}   \\

     \hline

\end{tabular}
\caption{Per-class mean accuracy (\%) on DomainNet.}
\label{table2}
\end{table*}

\subsection{Analysis}

\subsubsection{Different Degrees of Label Shift}

\begin{table}[t]
    \centering
    \begin{tabular}{@{}l@{}|@{\ \ }c@{\ \ } @{\ \ }c@{\ \ } @{\ \ }c@{\ \ } @{\ \ }c@{\ \ } @{\ \ }c@{\ \ } }

         \hline
         Imblanced Ratio & 1.00 & 0.20 & 0.10 & 0.05 & 0.02 \\

         \hline
         DAN &  61.7 & 60.7 & 57.2 & 53.5 & 47.0 \\

         DANN & 65.4 & 60.3 & 52.2 & 45.1 & 38.2 \\

         BSP & 68.2 & 63.0 & 55.5 & 49.2 & 41.4 \\

         CDAN & 70.0 & 64.6 & 57.8 & 48.5 & 38.9 \\

         MDD & 69.6 & 65.6 & 60.1 & 55.0 & 45.6 \\ 

         TIToK & 69.1 & 66.1 & 61.5 & 56.5 & 48.0 \\

         \hline
         centroIDA & \textbf{70.5} & \textbf{67.8} & \textbf{66.3} & \textbf{62.9} & \textbf{55.7} \\
         \hline
         
    \end{tabular}
    \caption{Per-class mean accuracy (\%) on the task Cl \hspace{-0.8mm}$\to$\hspace{-0.8mm} Pr with differen imbalanced ratios $p$. }
    \label{table3}
\end{table}

We design experiments with imbalanced ratios $p$ from $\{1.00, 0.20, 0.10, 0.05, 0.02\}$ to explore the influence of label shifts. Table \ref{table3} shows the results. Where in the case of $p=1.00$, the dataset has not been sampled and maintains its original distribution. According to the table results, it can be concluded that (\romannumeral1) as the degree of label shift increases, the performance of all methods deteriorates, indicating that label shifts have a negative impact on the model; (\romannumeral2) this negative impact is milder on centroIDA, indicating that centroIDA has a stronger ability to resist label shifts; and (\romannumeral3) centroIDA achieved the highest performance at all p-values, indicating that it has greater advantages in solving covariate shifts and label shifts. In short, centroIDA can efficiently resolve IDA problems.

\subsubsection{Hyper-parameter Analysis}
\begin{figure}[H]
\centering
\includegraphics[width=0.9\columnwidth]{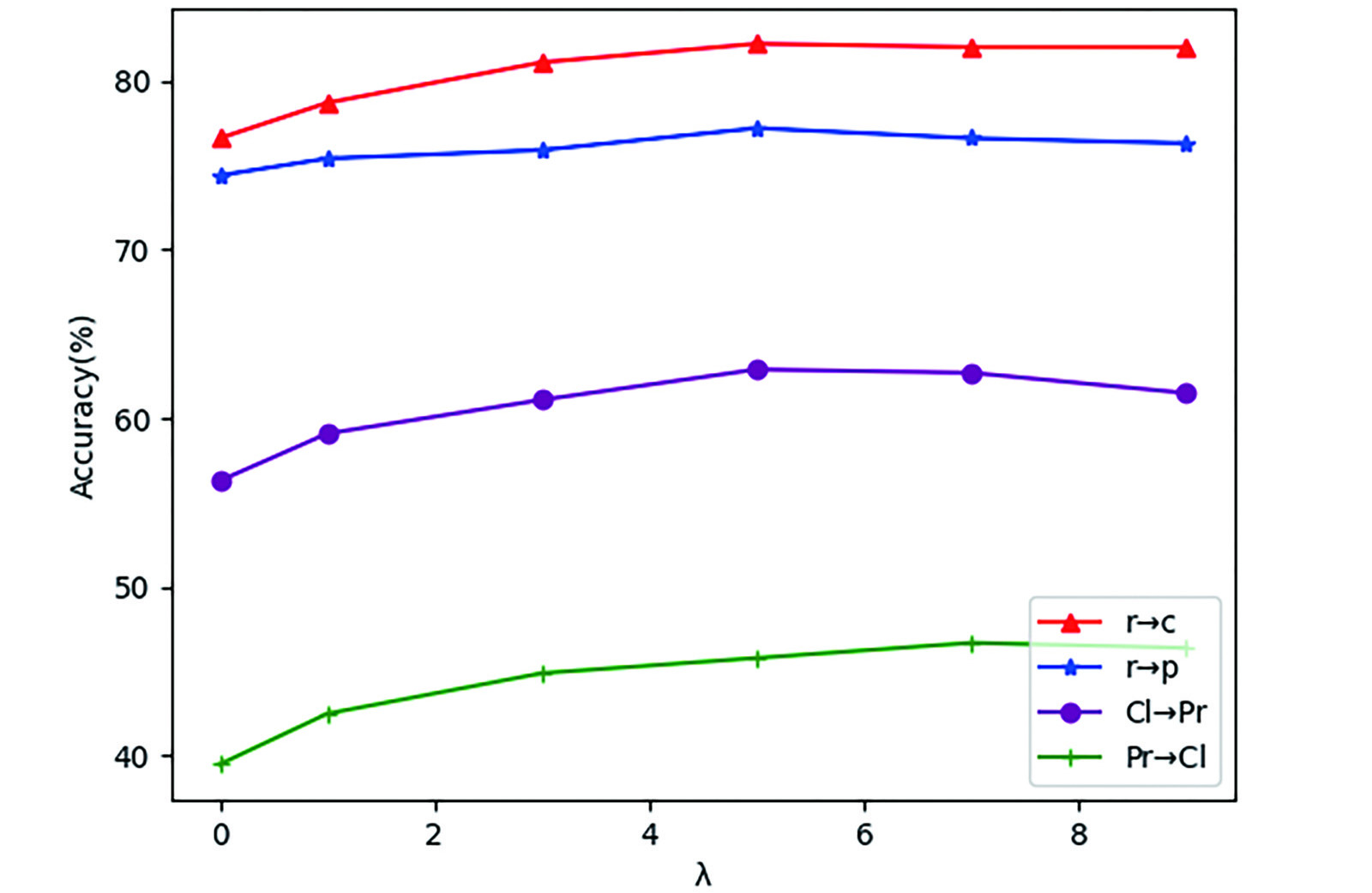} % Reduce the figure size so that it is slightly narrower than the column. Don't use precise values for figure width.This setup will avoid overfull boxes.
\caption{Performance of centroIDA at different value of $\lambda$.}
\label{fig5}
\end{figure}

Due to the existence of noise in the early phases of training, directly aligning according to Eq.(\ref{eq:(11)}) will result in more errors. Therefore, we set $\gamma$ to $\gamma=\frac{2}{\exp (-10\alpha)}-1$, where $\alpha$ varies linearly from $0$ to $1$ and set $\lambda$ to a constant value. In this section, we discussed the performance on four transfer tasks in the case of $\lambda$ from
$\{0,1,3,5,7,9\}$, as shown in Fig.\ref{fig5}. When $\lambda=0$, class-centroids alignment does not work, and the performance of the model is relatively poor. As the $\lambda$ increases, so does the focus on class-centroids alignment,  and the performance of all four tasks improved  and gradually stabilized. According to the results in Fig.\ref{fig5}, increasing the value of $\lambda$ when it is greater than $5$ does not result in any additional performance benefit. Therefore, in this article, we uniformly set $\lambda$ to $5$.

\subsubsection{Ablation Study}
\begin{table}[t]
    \centering
    \begin{tabular}{@{}l@{}| @{\ }c@{\ } @{\ }c@{\ } @{\ \ }c@{\ \ } @{\ \ }c@{\ \ } @{\ \ }c@{\ \ } }

         \hline
         \centering Methods & Cl \hspace{-0.8mm}$\to$\hspace{-0.8mm} Pr & Pr \hspace{-0.8mm}$\to$\hspace{-0.8mm} Ar & r \hspace{-0.8mm}$\to$\hspace{-0.8mm} c & r \hspace{-0.8mm}$\to$\hspace{-0.8mm} p & p \hspace{-0.8mm}$\to$\hspace{-0.8mm} c \\

         \hline
         Source Only & 43.0 & 42.5 & 60.2 & 65.1 & 55.9 \\
         \hline
         rm-Resample & 48.1 & 42.7 & 77.3 & 73.7 & 66.3\\
         
         rm-$loss\_c$ & 56.3 &50.5 & 76.6 & 74.4 & 73.7\\

         rm-$loss\_d$ &60.5 & 56.1 & 77.3 & 73.4 & 75.9\\

         centroIDA & \textbf{62.9} & \textbf{57.3} & \textbf{82.2} & \textbf{77.2} & \textbf{78.3}\\
         \hline
         
    \end{tabular}
    \caption{Ablation study. Performance of centroIDA and its variants on five tranfer tasks.}
    \label{table4}
\end{table}
We remove one of centroIDA's three core components and obtained its variants, respectively, to investigate the contribution of each component to the model. Table \ref{table4} shows that deleting any component from centroIDA will result in poor model performance, particularly removing the class-balanced re-sampling strategy. This is due to the fact that the calculation of accumulative class-centroids are rely on the class-balanced re-sampling strategy. Because of the data imbalance, minority classes have less opportunities to engage in training, making its accumulative class-centroids harder to update and unreliable. The class-balanced re-sampling strategy ensures not only the dependability of pseudo labels, but also the fairness of each class being sampled, so making each class centroid reliable.

\subsubsection{Complexity Analysis}  
The complexity of centroIDA is $\mathcal{O}(n^2)$, while the complexity of those methods based on prototype network are $\mathcal{O}(n^k+N)$, where $n$ is the batch size, the value of k depends on the algorithm strategy used in those methods, and $N$ ($0\le N \le N_{all}$, $N_{all}$ represents all samples numbers of a dataset) is a variable value representing the number of samples selected in the target domain. Due to the influence of the selection strategy on the value of N, there exists $N \gg n$ during the training. Therefore, centroIDA has lower algorithm complexity to achieve conditional feature alignment than that methods based on prototype network.

\section{Conclusion}
In this paper, a new method, centroIDA, is proposed to overcome the IDA problem. The method is based on class-balanced re-sampling strategy, accumulative class-centroids alignment and class-wise feature alignment. The accumulative class-centroids is originally designed and takes both covariate shifts and label shifts into account. Specifically, a trial experiment is firstly carried out to illustrate the limitations of existing UDA models in the IDA scenario. Then, the drawbacks of current IDA approaches are investigated, and the details of the centroIDA are introduced. Finally, a series of experiments have been conducted and verify the centroIDA's state-of-the-art performance. We believe that our approach will encourage more attention to both label shift and covariate shift in IDA, allowing domain adaptation to tackle more practical challenges.

\bibliography{centroIDA}

\begin{thebibliography}{49}
\providecommand{\natexlab}[1]{#1}

\bibitem[{Alexandari, Kundaje, and Shrikumar(2020)}]{alexandari2020maximum}
Alexandari, A.; Kundaje, A.; and Shrikumar, A. 2020.
\newblock Maximum likelihood with bias-corrected calibration is hard-to-beat at
  label shift adaptation.
\newblock In \emph{International Conference on Machine Learning}, 222--232.
  PMLR.

\bibitem[{Azizzadenesheli(2022)}]{9445225}
Azizzadenesheli, K. 2022.
\newblock Importance Weight Estimation and Generalization in Domain Adaptation
  Under Label Shift.
\newblock \emph{IEEE Transactions on Pattern Analysis and Machine
  Intelligence}, 44(10): 6578--6584.

\bibitem[{Azizzadenesheli et~al.(2019)Azizzadenesheli, Liu, Yang, and
  Anandkumar}]{azizzadenesheli2019regularized}
Azizzadenesheli, K.; Liu, A.; Yang, F.; and Anandkumar, A. 2019.
\newblock Regularized learning for domain adaptation under label shifts.
\newblock \emph{arXiv preprint arXiv:1903.09734}.

\bibitem[{Cao et~al.(2018)Cao, Ma, Long, and Wang}]{cao2018partial}
Cao, Z.; Ma, L.; Long, M.; and Wang, J. 2018.
\newblock Partial adversarial domain adaptation.
\newblock In \emph{Proceedings of the European conference on computer vision
  (ECCV)}, 135--150.

\bibitem[{Chen et~al.(2019)Chen, Wang, Long, and
  Wang}]{chen2019transferability}
Chen, X.; Wang, S.; Long, M.; and Wang, J. 2019.
\newblock Transferability vs. discriminability: Batch spectral penalization for
  adversarial domain adaptation.
\newblock In \emph{International conference on machine learning}, 1081--1090.
  PMLR.

\bibitem[{Dong, Gong, and Zhu(2018)}]{dong2018imbalanced}
Dong, Q.; Gong, S.; and Zhu, X. 2018.
\newblock Imbalanced deep learning by minority class incremental rectification.
\newblock \emph{IEEE transactions on pattern analysis and machine
  intelligence}, 41(6): 1367--1381.

\bibitem[{Ganin et~al.(2016)Ganin, Ustinova, Ajakan, Germain, Larochelle,
  Laviolette, Marchand, and Lempitsky}]{ganin2016domain}
Ganin, Y.; Ustinova, E.; Ajakan, H.; Germain, P.; Larochelle, H.; Laviolette,
  F.; Marchand, M.; and Lempitsky, V. 2016.
\newblock Domain-adversarial training of neural networks.
\newblock \emph{The journal of machine learning research}, 17(1): 2096--2030.

\bibitem[{Guo et~al.(2017)Guo, Pleiss, Sun, and
  Weinberger}]{guo2017calibration}
Guo, C.; Pleiss, G.; Sun, Y.; and Weinberger, K.~Q. 2017.
\newblock On calibration of modern neural networks.
\newblock In \emph{International conference on machine learning}, 1321--1330.
  PMLR.

\bibitem[{Jiang et~al.(2020)Jiang, Lao, Matwin, and Havaei}]{jiang2020implicit}
Jiang, X.; Lao, Q.; Matwin, S.; and Havaei, M. 2020.
\newblock Implicit class-conditioned domain alignment for unsupervised domain
  adaptation.
\newblock In \emph{International Conference on Machine Learning}, 4816--4827.
  PMLR.

\bibitem[{Jin et~al.(2020)Jin, Wang, Long, and Wang}]{jin2020minimum}
Jin, Y.; Wang, X.; Long, M.; and Wang, J. 2020.
\newblock Minimum class confusion for versatile domain adaptation.
\newblock In \emph{Computer Vision--ECCV 2020: 16th European Conference,
  Glasgow, UK, August 23--28, 2020, Proceedings, Part XXI 16}, 464--480.
  Springer.

\bibitem[{Jing, Xu, and Ding(2021)}]{9547057}
Jing, T.; Xu, B.; and Ding, Z. 2021.
\newblock Towards Fair Knowledge Transfer for Imbalanced Domain Adaptation.
\newblock \emph{IEEE Transactions on Image Processing}, 30: 8200--8211.

\bibitem[{JT and Yang(2011)}]{jt2011domain}
JT, P. S. T. I.~K.; and Yang, Q. 2011.
\newblock Domain adaptation via transfer component analysis.
\newblock \emph{IEEE Trans Neural Netw}, 22(2): 199.

\bibitem[{Kang et~al.(2019{\natexlab{a}})Kang, Xie, Rohrbach, Yan, Gordo, Feng,
  and Kalantidis}]{kang2019decoupling}
Kang, B.; Xie, S.; Rohrbach, M.; Yan, Z.; Gordo, A.; Feng, J.; and Kalantidis,
  Y. 2019{\natexlab{a}}.
\newblock Decoupling representation and classifier for long-tailed recognition.
\newblock \emph{arXiv preprint arXiv:1910.09217}.

\bibitem[{Kang et~al.(2019{\natexlab{b}})Kang, Jiang, Yang, and
  Hauptmann}]{kang2019contrastive}
Kang, G.; Jiang, L.; Yang, Y.; and Hauptmann, A.~G. 2019{\natexlab{b}}.
\newblock Contrastive adaptation network for unsupervised domain adaptation.
\newblock In \emph{Proceedings of the IEEE/CVF conference on computer vision
  and pattern recognition}, 4893--4902.

\bibitem[{Kouw and Loog(2018)}]{kouw2018introduction}
Kouw, W.~M.; and Loog, M. 2018.
\newblock An introduction to domain adaptation and transfer learning.
\newblock \emph{arXiv preprint arXiv:1812.11806}.

\bibitem[{Lipton, Wang, and Smola(2018)}]{lipton2018detecting}
Lipton, Z.; Wang, Y.-X.; and Smola, A. 2018.
\newblock Detecting and correcting for label shift with black box predictors.
\newblock In \emph{International conference on machine learning}, 3122--3130.
  PMLR.

\bibitem[{Liu et~al.(2019)Liu, Long, Wang, and Jordan}]{liu2019transferable}
Liu, H.; Long, M.; Wang, J.; and Jordan, M. 2019.
\newblock Transferable adversarial training: A general approach to adapting
  deep classifiers.
\newblock In \emph{International Conference on Machine Learning}, 4013--4022.
  PMLR.

\bibitem[{Liu et~al.(2021)Liu, Guo, Li, Xing, You, Kuo, El~Fakhri, and
  Woo}]{liu2021adversarial}
Liu, X.; Guo, Z.; Li, S.; Xing, F.; You, J.; Kuo, C.-C.~J.; El~Fakhri, G.; and
  Woo, J. 2021.
\newblock Adversarial unsupervised domain adaptation with conditional and label
  shift: Infer, align and iterate.
\newblock In \emph{Proceedings of the IEEE/CVF international conference on
  computer vision}, 10367--10376.

\bibitem[{Liu et~al.(2020)Liu, Han, Bai, Ge, Wang, Han, Li, You, and
  Lu}]{liu2020importance}
Liu, X.; Han, Y.; Bai, S.; Ge, Y.; Wang, T.; Han, X.; Li, S.; You, J.; and Lu,
  J. 2020.
\newblock Importance-aware semantic segmentation in self-driving with discrete
  wasserstein training.
\newblock In \emph{Proceedings of the AAAI Conference on Artificial
  Intelligence}, volume~34, 11629--11636.

\bibitem[{Long et~al.(2015)Long, Cao, Wang, and Jordan}]{long2015learning}
Long, M.; Cao, Y.; Wang, J.; and Jordan, M. 2015.
\newblock Learning transferable features with deep adaptation networks.
\newblock In \emph{International conference on machine learning}, 97--105.
  PMLR.

\bibitem[{Long et~al.(2018)Long, Cao, Wang, and Jordan}]{long2018conditional}
Long, M.; Cao, Z.; Wang, J.; and Jordan, M.~I. 2018.
\newblock Conditional adversarial domain adaptation.
\newblock \emph{Advances in neural information processing systems}, 31.

\bibitem[{Long et~al.(2017)Long, Zhu, Wang, and Jordan}]{long2017deep}
Long, M.; Zhu, H.; Wang, J.; and Jordan, M.~I. 2017.
\newblock Deep transfer learning with joint adaptation networks.
\newblock In \emph{International conference on machine learning}, 2208--2217.
  PMLR.

\bibitem[{Mei et~al.(2020)Mei, Zhu, Zou, and Zhang}]{mei2020instance}
Mei, K.; Zhu, C.; Zou, J.; and Zhang, S. 2020.
\newblock Instance adaptive self-training for unsupervised domain adaptation.
\newblock In \emph{Computer Vision--ECCV 2020: 16th European Conference,
  Glasgow, UK, August 23--28, 2020, Proceedings, Part XXVI 16}, 415--430.
  Springer.

\bibitem[{Morerio, Cavazza, and Murino(2017)}]{morerio2017minimal}
Morerio, P.; Cavazza, J.; and Murino, V. 2017.
\newblock Minimal-entropy correlation alignment for unsupervised deep domain
  adaptation.
\newblock \emph{arXiv preprint arXiv:1711.10288}.

\bibitem[{Pan et~al.(2019)Pan, Yao, Li, Wang, Ngo, and
  Mei}]{pan2019transferrable}
Pan, Y.; Yao, T.; Li, Y.; Wang, Y.; Ngo, C.-W.; and Mei, T. 2019.
\newblock Transferrable prototypical networks for unsupervised domain
  adaptation.
\newblock In \emph{Proceedings of the IEEE/CVF conference on computer vision
  and pattern recognition}, 2239--2247.

\bibitem[{Panareda~Busto and Gall(2017)}]{panareda2017open}
Panareda~Busto, P.; and Gall, J. 2017.
\newblock Open set domain adaptation.
\newblock In \emph{Proceedings of the IEEE international conference on computer
  vision}, 754--763.

\bibitem[{Pei et~al.(2018)Pei, Cao, Long, and Wang}]{pei2018multi}
Pei, Z.; Cao, Z.; Long, M.; and Wang, J. 2018.
\newblock Multi-adversarial domain adaptation.
\newblock In \emph{Proceedings of the AAAI conference on artificial
  intelligence}, volume~32.

\bibitem[{Peng et~al.(2019)Peng, Bai, Xia, Huang, Saenko, and
  Wang}]{peng2019moment}
Peng, X.; Bai, Q.; Xia, X.; Huang, Z.; Saenko, K.; and Wang, B. 2019.
\newblock Moment matching for multi-source domain adaptation.
\newblock In \emph{Proceedings of the IEEE/CVF international conference on
  computer vision}, 1406--1415.

\bibitem[{Prabhu et~al.(2021)Prabhu, Khare, Kartik, and
  Hoffman}]{prabhu2021sentry}
Prabhu, V.; Khare, S.; Kartik, D.; and Hoffman, J. 2021.
\newblock Sentry: Selective entropy optimization via committee consistency for
  unsupervised domain adaptation.
\newblock In \emph{Proceedings of the IEEE/CVF International Conference on
  Computer Vision}, 8558--8567.

\bibitem[{Ren et~al.(2018)Ren, Zeng, Yang, and Urtasun}]{ren2018learning}
Ren, M.; Zeng, W.; Yang, B.; and Urtasun, R. 2018.
\newblock Learning to reweight examples for robust deep learning.
\newblock In \emph{International conference on machine learning}, 4334--4343.
  PMLR.

\bibitem[{Shin et~al.(2020)Shin, Woo, Pan, and Kweon}]{shin2020two}
Shin, I.; Woo, S.; Pan, F.; and Kweon, I.~S. 2020.
\newblock Two-phase pseudo label densification for self-training based domain
  adaptation.
\newblock In \emph{Computer Vision--ECCV 2020: 16th European Conference,
  Glasgow, UK, August 23--28, 2020, Proceedings, Part XIII 16}, 532--548.
  Springer.

\bibitem[{Snell, Swersky, and Zemel(2017)}]{snell2017prototypical}
Snell, J.; Swersky, K.; and Zemel, R. 2017.
\newblock Prototypical networks for few-shot learning.
\newblock \emph{Advances in neural information processing systems}, 30.

\bibitem[{Sun, Feng, and Saenko(2016)}]{sun2016return}
Sun, B.; Feng, J.; and Saenko, K. 2016.
\newblock Return of frustratingly easy domain adaptation.
\newblock In \emph{Proceedings of the AAAI conference on artificial
  intelligence}, volume~30.

\bibitem[{Sun and Saenko(2016)}]{sun2016deep}
Sun, B.; and Saenko, K. 2016.
\newblock Deep coral: Correlation alignment for deep domain adaptation.
\newblock In \emph{Computer Vision--ECCV 2016 Workshops: Amsterdam, The
  Netherlands, October 8-10 and 15-16, 2016, Proceedings, Part III 14},
  443--450. Springer.

\bibitem[{Tan, Peng, and Saenko(2020)}]{tan2020class}
Tan, S.; Peng, X.; and Saenko, K. 2020.
\newblock Class-imbalanced domain adaptation: an empirical odyssey.
\newblock In \emph{Computer Vision--ECCV 2020 Workshops: Glasgow, UK, August
  23--28, 2020, Proceedings, Part I 16}, 585--602. Springer.

\bibitem[{Tian et~al.(2020{\natexlab{a}})Tian, Tang, Hu, Ren, and
  Zhang}]{9234105}
Tian, L.; Tang, Y.; Hu, L.; Ren, Z.; and Zhang, W. 2020{\natexlab{a}}.
\newblock Domain Adaptation by Class Centroid Matching and Local Manifold
  Self-Learning.
\newblock \emph{IEEE Transactions on Image Processing}, 29: 9703--9718.

\bibitem[{Tian et~al.(2020{\natexlab{b}})Tian, Tang, Hu, Ren, and
  Zhang}]{tian2020domain}
Tian, L.; Tang, Y.; Hu, L.; Ren, Z.; and Zhang, W. 2020{\natexlab{b}}.
\newblock Domain adaptation by class centroid matching and local manifold
  self-learning.
\newblock \emph{IEEE Transactions on Image Processing}, 29: 9703--9718.

\bibitem[{Tzeng et~al.(2017)Tzeng, Hoffman, Saenko, and
  Darrell}]{tzeng2017adversarial}
Tzeng, E.; Hoffman, J.; Saenko, K.; and Darrell, T. 2017.
\newblock Adversarial discriminative domain adaptation.
\newblock In \emph{Proceedings of the IEEE conference on computer vision and
  pattern recognition}, 7167--7176.

\bibitem[{Venkateswara et~al.(2017)Venkateswara, Eusebio, Chakraborty, and
  Panchanathan}]{venkateswara2017deep}
Venkateswara, H.; Eusebio, J.; Chakraborty, S.; and Panchanathan, S. 2017.
\newblock Deep hashing network for unsupervised domain adaptation.
\newblock In \emph{Proceedings of the IEEE conference on computer vision and
  pattern recognition}, 5018--5027.

\bibitem[{Wang et~al.(2023)Wang, Chen, Liu, Li, and Chen}]{wang2023titok}
Wang, Y.; Chen, Q.; Liu, Y.; Li, W.; and Chen, S. 2023.
\newblock TIToK: A solution for bi-imbalanced unsupervised domain adaptation.
\newblock \emph{Neural Networks}, 164: 81--90.

\bibitem[{Wang et~al.(2017)Wang, Li, Dai, and Van~Gool}]{wang2017deep}
Wang, Y.; Li, W.; Dai, D.; and Van~Gool, L. 2017.
\newblock Deep domain adaptation by geodesic distance minimization.
\newblock In \emph{Proceedings of the IEEE International Conference on Computer
  Vision Workshops}, 2651--2657.

\bibitem[{Wei et~al.(2020)Wei, Shen, Chen, and Ma}]{wei2020theoretical}
Wei, C.; Shen, K.; Chen, Y.; and Ma, T. 2020.
\newblock Theoretical analysis of self-training with deep networks on unlabeled
  data.
\newblock \emph{arXiv preprint arXiv:2010.03622}.

\bibitem[{Wu et~al.(2019)Wu, Winston, Kaushik, and Lipton}]{wu2019domain}
Wu, Y.; Winston, E.; Kaushik, D.; and Lipton, Z. 2019.
\newblock Domain adaptation with asymmetrically-relaxed distribution alignment.
\newblock In \emph{International conference on machine learning}, 6872--6881.
  PMLR.

\bibitem[{Yang et~al.(2021)Yang, Yang, Wang, Cao, Zou, and
  Xie}]{yang2021advancing}
Yang, J.; Yang, J.; Wang, S.; Cao, S.; Zou, H.; and Xie, L. 2021.
\newblock Advancing imbalanced domain adaptation: Cluster-level discrepancy
  minimization with a comprehensive benchmark.
\newblock \emph{IEEE Transactions on Cybernetics}.

\bibitem[{Yang et~al.(2020{\natexlab{a}})Yang, Zou, Zhou, and
  Xie}]{yang2020towards}
Yang, J.; Zou, H.; Zhou, Y.; and Xie, L. 2020{\natexlab{a}}.
\newblock Towards stable and comprehensive domain alignment: Max-margin
  domain-adversarial training.
\newblock \emph{arXiv preprint arXiv:2003.13249}.

\bibitem[{Yang et~al.(2020{\natexlab{b}})Yang, Deng, Liu, and
  Tao}]{yang2020heterogeneous}
Yang, X.; Deng, C.; Liu, T.; and Tao, D. 2020{\natexlab{b}}.
\newblock Heterogeneous graph attention network for unsupervised
  multiple-target domain adaptation.
\newblock \emph{IEEE Transactions on Pattern Analysis and Machine
  Intelligence}, 44(4): 1992--2003.

\bibitem[{Zhang et~al.(2019)Zhang, Liu, Long, and Jordan}]{zhang2019bridging}
Zhang, Y.; Liu, T.; Long, M.; and Jordan, M. 2019.
\newblock Bridging theory and algorithm for domain adaptation.
\newblock In \emph{International conference on machine learning}, 7404--7413.
  PMLR.

\bibitem[{Zhang et~al.(2018)Zhang, Wang, Cai, and Song}]{zhang2018unsupervised}
Zhang, Y.; Wang, N.; Cai, S.; and Song, L. 2018.
\newblock Unsupervised domain adaptation by mapped correlation alignment.
\newblock \emph{IEEE Access}, 6: 44698--44706.

\bibitem[{Zou et~al.(2019)Zou, Zhou, Yang, Liu, Das, and
  Spanos}]{zou2019consensus}
Zou, H.; Zhou, Y.; Yang, J.; Liu, H.; Das, H.~P.; and Spanos, C.~J. 2019.
\newblock Consensus adversarial domain adaptation.
\newblock In \emph{Proceedings of the AAAI conference on artificial
  intelligence}, volume~33, 5997--6004.

\end{thebibliography}

\end{document}